# Segmentation of Offline Handwritten Bengali Script

S. Basu [+], C. Chaudhuri[*], M. Kundu[*],
M. Nasipuri[*], D.K. Basu[*]

[+] Computer Sc. & Engg. Dept., MCKV Institute of Engineering, Liluah, Howrah-711204, India
[*] Computer Sc. & Engg. Dept., Jadavpur University, Kolkata-700032, India

**Abstract:** Character segmentation has long been one of the most critical areas of optical character recognition process. Through this operation, an image of a sequence of characters, which may be connected in some cases, is decomposed into sub-images of individual alphabetic symbols. In this paper, segmentation of cursive handwritten script of world's fourth popular language, Bengali, is considered. Unlike English script, Bengali handwritten characters and its components often encircle the main character, making the conventional segmentation methodologies inapplicable. Experimental results, using the proposed segmentation technique, on sample cursive handwritten data containing 218 ideal segmentation points show a success rate of 97.7%. Further feature-analysis on these segments may lead to actual recognition of handwritten cursive Bengali script.

## 1. INTRODUCTION

Character segmentation is one of the most important decision processes for optical character recognition (OCR). Isolating individual alphabetic characters in the script image is often significant enough to make a decisive contribution towards the success rate of the overall system.

An OCR system may be designed to work for either of on-line and off-line purposes. On-line OCR systems collect input data by recording the order of strokes made by the write on an electronic bit-pad, and off-line OCR systems do the same by recording the pixel by pixel digital image of the entire writing with a digital scanner. OCR has a wide field of application covering handwritten document transcription, automatic mail address recognition, machine processing of bank-checks, faxes etc.

Off-line OCR of hand written words has long been an active area research. Some important contributions so far made in this field involve analysis of English texts [1], [2], [3], [5], Chinese script [6] and Arabic characters [9]. With this background of research, the present work considers Bengali script for developing suitable techniques for off-line OCR with it. About 200 million people of Eastern India and Bangladesh use Bengali script for communication. So there is an urgent need for development of OCR systems capable of handling handwritten Bengali script. The OCR involving printed Bengali script has already been addressed in [8]. But problems related to OCR of hand written Bengali script still constitute an unexplored area of research.

Segmentation and subsequent analysis of handwritten Bengali script is a challenging pattern recognition problem. Firstly, the vocabulary is very large. The modern Bengali script alphabet consists of 11 vowels, 39 consonants (shown in fig 1a and fig 1b). Secondly, Bengali characters are very complex. Thirdly, many Bengali characters look very similar to each other and there are great pattern variations between samples of the same character written by the same person.

Most characters in this script have horizontal lines at the top called head-lines. In cursive writing, these head-lines mostly join with one another in a word and the word appears as a single component hanging from the head line. In Bengali script a vowel followed by a consonant takes a modified shape (shown in fig 1c), which is placed at the left, right, top, bottom (or their combinations) of the consonant depending on the vowel. These modified shapes often overlap and encircle the original character. If other shapes or characters combine with the original character, then they are called compound characters. A consonant or vowel following a consonant often takes a compound orthographic shape, which may be termed as compound character. There are 10 modified shapes and around 250 compound characters in Bengali script [4].

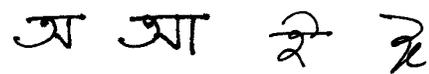
Fig. 1a. Sample Bengali vowels

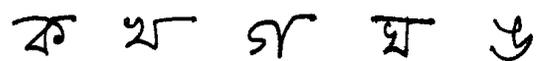
Fig. 1b. Sample Bengali consonants

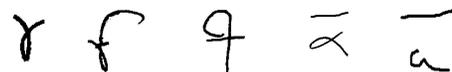
Fig. 1c. Sample modified shapes of Bengali script

In this paper, a hybrid model of image based dissection and recognition based segmentation is proposed. Through image based dissection approach, segments are identified based on 'character-like' properties and are cut into meaningful components, whereas in recognition based segmentation approach, the system searches the image components based on image features that match classes of its pattern [1].

The system and the approach that we propose in this paper has several favorable practical advantages. It is designed to work with most general handwriting styles with mixed combinations of discrete and cursive handwritten Bengali scripts. The overall design is modular (shown in fig. 2) and each module can be fine tuned or replaced with improved versions without major impact on other modules.

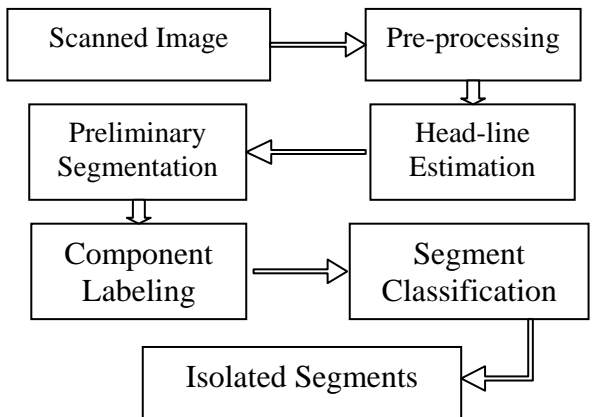

Fig.2. Schematic of the segmentation process

## 2. THE SEGMENTATION PROCESS

To capture data from a handwritten document, a conventional flatbed scanner of resolution 300 dpi is used. In the database collected for this research, words were written on plain, white sheet of paper with black pen, which gave clear stroke with sharp edges. The collected data is assumed to be slope corrected.

As mentioned earlier, the proposed process of segmentation consists of several structured modules. Following subsections discuss these modules in details.

### 2.1. Pre-processing

Several pre-processing steps are done on the composite raw image before actual segmentation process starts. Firstly, the input scanned image is converted into its corresponding binary equivalent data. Following sub-modules subsequently attempt to classify the input data.

#### 2.1.1. Word Segmentation

The basic objective of this step is to segment each scanned image page into isolated words. A single scanned image may contain several lines of data and each line subsequently contains several disjoint words. The basic word segmentation algorithm scans the input image horizontally; generates a horizontal pixel density histogram for the input data. Comparing the row-wise intensity between successive rows of the histogram with a predefined threshold $k_1$, individual lines are isolated (shown in fig. 3a).

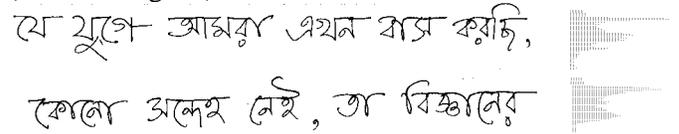

Fig. 3a. Line segmentation using row histograms

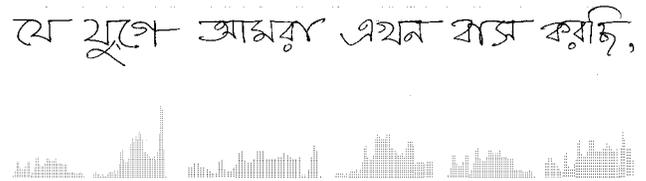

Fig. 3b. Word segmentation using column

Each isolated line is then scanned vertically and a vertical pixel density histogram is generated. By comparing the column-wise intensity of the histogram with a predefine threshold $k_2$, word segmentation boundaries are identified (shown in fig. 3b).

#### 2.1.2. Format Analysis

Each of the isolated words is subdivided into four horizontal imaginary regions (shown in fig.4). Selection of these regions is based on the basic characteristics of Bengali script. Experimental analysis reveals that in most of the words in Bengali script, character components or modified shapes extend above the head-line or lie below the actual characters. The components that extend over head-line are called ascendants and the character components or modified shapes that lie below the actual character are called descendents. Ascendants normally lie in region-1, whereas the descendents generally lie in region-4 and the main character body is dominated in region-2 and region-3.

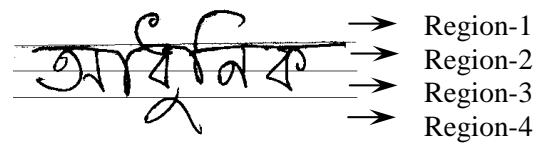

Fig.4. Format analysis of a sample handwritten word

The format analysis algorithm divides the input word data into four equal-width regions. Line segmentation threshold $k_1$ is chosen carefully to classify between region-1 and region-4 of successive lines. In the proposed segmentation process, the presences of character contours in different regions are considered to be the most important guideline factors for handwritten character segmentation. This format analysis phase acts as a building block for the subsequent steps like head-line estimation and preliminary segmentation.

### 2.2. Head-line Estimation

In Bengali script, head-line, common in most of the characters, is considered to be the most prominent feature element. The character occurrence statistics in Bengali script confirms that out of twelve most frequent characters, only one character has no head-line [4]. So, it is likely that most Bengali words will have head-line too. In continuous writing, these characters are often connected through head-lines and appear as a single component in a word.

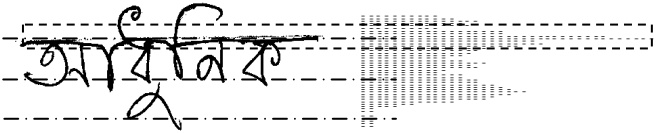

Fig. 5. Approximate head-line estimation

Proper identification of this head-line is a challenging task for character segmentation. Several modified shapes often encircle the main character and the head-line. Format analysis on the input word data shows that the probability of occurrence of the head-line is highest along the boundary of region-1 and region-2. The head-line estimation algorithm generates a horizontal histogram for the input word data. Maximum pixel density row within region-1 and region-2 is identified. This row is identified as the most significant headline row, having pixel intensity equals to MAX (say). If the other neighbouring rows of MAX have frequency greater than $w * MAX$ ($w$ is a predefined experimental constant and $0 <= w <= 1$), then those neighbouring rows are considered to be the head-line rows of the input data word (shown in fig. 5).

### 2.3. Preliminary Segmentation

In this module, segments are identified based on 'character-like' features. The process of cutting up the image into meaningful components is given the name 'dissection'. Each word is analyzed with the following features to identify preliminary segmentation points.

Table 1 shows the few important features used to compute the segmentation weightage along the head-line. If the calculated segmentation weightage of a point along the head-line is greater than a pre-defined threshold $\delta$, then the columnar strip along the head-line is considered to be a preliminary segmentation point.

| Feature ID | Feature Description |
|---|---|
| 1 | Vertical Density Histogram of image through Region-2 and Region-3 |
| 2 | Vertical Density Histogram of image through Region-1 |
| 3 | Vertical Area Histogram of image through region-2 and region-3 |
| 4 | Vertical Thickness ratio of the head-line along the head-line region |
| 5 | Vertical Area Histogram of the head-line along the head-line region |
| 6 | Connectivity of Region-1 and Region-2 with head-line |

Table 1. List of Segmentation features

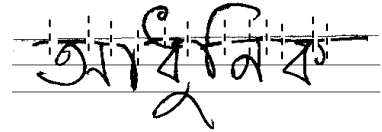

Fig. 6. Identified Preliminary Segmentation points

Once the preliminary segmentation points are identified, the input image is segmented along the head-line region through vertical strips depending on their segmentation weightage (shown in fig. 6).

### 2.3. Component Labeling

Preliminary segmentation points disjoin the characters, character components and modified shapes along the head-line. But these components often encircle each other making the process of isolating the components difficult. The connected component-labeling algorithm [8] is implemented to isolate each preliminary segment (as shown in fig 7).

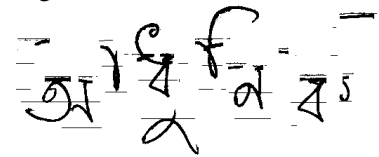

Fig. 7. Preliminary segments being isolated through connected component labeling

### 2.4. Segment Classification

Each of the isolated segments may not necessarily represent a character or a modified shape. In Bengali character set, few characters (classified in fig.8), may generate disjoint character components through preliminary segmentation process. These characters get segmented within them. Proper classification and re-unification of these components using recognition based segmentation is another challenging task for segmentation of handwritten Bengali script

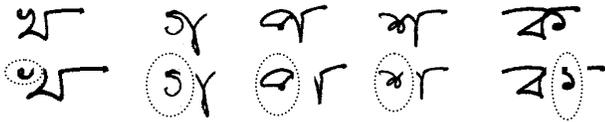

Fig.8. Sample characters with isolated broken segments

## 3. EXPERIMENTAL RESULTS

The purpose of these experiments is to test the potential of the proposed segmentation process for handwritten Bengali script. For the experiment, several handwritten text documents were collected. Samples taken from ten different persons have been analyzed. Both cursive and isolated text inputs were considered.

| Input Data | After Preliminary Segmentation | Final Segmented Output |
|---|---|---|
| আমরুণ | আমরুণ | আমরুণ |
| বিক্ষানের | বিক্ষানের | বিক্ষানের |
| আধুনিক | আধুনিক | আধুনিক |

Fig. 9. Segmentation results on some sample handwritten data

The proposed algorithm was tested on a sample sheet of paper containing 218 ideal segmentation points. The inputs were so chosen, that they roughly represent the variety of the input data. Word segmentation and head-line estimation modules produced 100% success results on the test data. Sample results on preliminary segmentation and final classification are shown in fig.9. The overall segmentation process fares a 97.7% success rate on the sample data.

## 4. CONCLUSION

In this paper, several component modules and a system organization for offline handwritten Bengali script segmentation process is described. The experiment has demonstrated the system's ability to identify majority of the segmentation points of the handwritten words. Future research works may concentrate on the analysis of recognition based segmentation approach. The experiment excluded the compound characters available in Bengali script. Extensive analysis on these compound characters may lead to an overall improvement in the generality of the process. Future work may also include identification of overlapping pen-strokes with an improvement of all the existing modules.

## ACKNOWLEDGEMENT

Authors are thankful to the "Center for Microprocessor Application for Training Education and Research", "Project on Storage Retrieval and Understanding of Video for Multimedia" and Computer Science & Engineering Department, Jadavpur University, for providing infrastructure facilities during progress of the work. One of the authors, Mr. S. Basu, is thankful to MCKV Institute of Engineering for kindly permitting to carry on the research work.